# Multilingual Coreference Resolution in Low-resource South Asian Languages

**Ritwik Mishra**[1], **Pooja Desur**[2], **Rajiv Ratn Shah**[1], **Ponnurangam Kumaraguru**[2]
[1]Indraprastha Institute of Information Technology, Delhi
[2]International Institute of Information Technology, Hyderabad
ritwikm@iiitd.ac.in, pooja.desur@alumni.iiit.ac.in, rajivratn@iiitd.ac.in, pk.guru@iiit.ac.in

## Abstract

Coreference resolution involves the task of identifying text spans within a discourse that pertain to the same real-world entity. While this task has been extensively explored in the English language, there has been a notable scarcity of publicly accessible resources and models for coreference resolution in South Asian languages. We introduce a Translated dataset for Multilingual Coreference Resolution (TransMuCoRes) in 31 South Asian languages using off-the-shelf tools for translation and word-alignment. Nearly all of the predicted translations successfully pass a sanity check, and 75% of English references align with their predicted translations. Using multilingual encoders, two off-the-shelf coreference resolution models were trained on a concatenation of TransMuCoRes and a Hindi coreference resolution dataset with manual annotations. The best performing model achieved a score of 64 and 68 for LEA F1 and CoNLL F1, respectively, on our test-split of Hindi golden set. This study is the first to evaluate an end-to-end coreference resolution model on a Hindi golden set. Furthermore, this work underscores the limitations of current coreference evaluation metrics when applied to datasets with split antecedents, advocating for the development of more suitable evaluation metrics.

## 1. Introduction

The phenomenon of referring to an expression previously mentioned in a discourse is widespread in natural languages. In the written text, it circumvents the repetition of expressions and engenders a sequence of coherent and interconnected sentences. For instance, consider the paragraph: "***John** is a good student. **He** asks intelligent questions and helps others. No wonder everybody loves **the boy**.*" These sentences are interconnected, as various referring expressions (highlighted in bold) are used to refer to the same entity named "*John*". Coreference resolution is an automated process that identifies referring expressions in a given text and locates the closest expression to which it refers. It acts as a useful preprocessing step and helps in many downstream tasks like entity linking (Kundu et al., 2018), Question-Answering (QA) (Bhattacharjee et al., 2020), and chatbots (Zhu et al., 2018).

Several end-to-end coreference resolution tools are currently available for English (Dobrovolskii, 2021), Arabic (Aloraini et al., 2020), and various European languages (David, 2022). However, to the best of our knowledge, no such tool is available for coreference resolution in any South Asian language, despite the presence of multiple works in this field (Sikdar et al., 2016; Senapati and Garain, 2013; Khandale and Mahender, 2019a; Ram and Devi, 2017). Our study is specifically focused on South Asian languages as they are native to approximately 25% of the global population, and three of the ten most widely spoken languages worldwide hail from this region (Ethnologue, 2021). Consequently, the main contributions of our work are as follows:

1. We introduce[1] a **Trans**lated dataset for **Mu**ltilingual **Co**reference **Res**olution (TransMuCoRes)[2] in 31 South Asian languages.

2. We release checkpoints for two off-the-shelf coreference resolution models that have been fine-tuned on TransMuCoRes dataset and the manually annotated Hindi coreference resolution dataset by Mujadia et al. (2016).

3. We also highlight the limitations in current

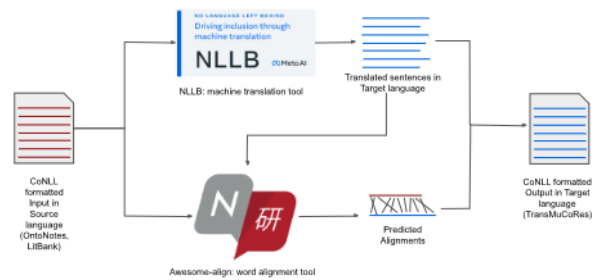

Figure 1: Overall pipeline used to construct the **Trans**lated dataset for **Mu**ltilingual **Co**reference **Res**olution (TransMuCoRes).

---

[1]https://github.com/ritwikmishra/transmucores
[2]pronounced *trans-mew-cores*

| Language | Assamese | Awadhi | Bengali | Bhojpuri | Tibetan |
|---|---|---|---|---|---|
| Script | Bangla | Devanagri | Bangla | Devanagri | Uchen |
| FLORES-200 code | asm_Beng | awa_Deva | ben_Beng | bho_Deva | bod_Tibt |
| Language | Dzongkha | Gujarati | Hindi | Chhattisgarhi | Kannada |
| Script | Uchen | Gujarati | Devanagri | Devanagri | Kannada |
| FLORES-200 code | dzo_Tibt | guj_Gujr | hin_Deva | hne_Deva | kan_Knda |
| Language | Kashmiri | Magihi | Maithili | Malayalam | Marathi |
| Script | Arabic | Devanagari | Devanagari | Malayalam | Devanagri |
| FLORES-200 code | kas_Arab | mag_Deva | mai_Deva | mal_Mlym | mar_Deva |
| Language | Meitei | Burmese | Nepali | Odia | Punjabi |
| Script | Bangla | Burmese | Devanagri | Kalinga | Gurumukhi |
| FLORES-200 code | mni_Beng | mya_Mymr | npi_Deva | ory_Orya | pan_Guru |
| Language | Pashto | Persian | Santali | Sinhala | Sindhi |
| Script | Arabic | Arabic | Ol Chiki | Sinhala | Arabic |
| FLORES-200 code | pbt_Arab | prs_Arab | sat_Beng | sin_Sinh | snd_Arab |
| Language | Tamil | Telugu | Tajik | Uyghur | Urdu |
| Script | Tamil | Telugu | Cyrillic | Arabic | Arabic |
| FLORES-200 code | tam_Taml | tel_Telu | tgk_Cyrl | uig_Arab | urd_Arab |
| Language | Uzbek | | | | |
| Script | Latin | | | | |
| FLORES-200 code | uzn_Latn | | | | |

Table 1: A catalogue of South Asian languages supported by TransMuCoRes. Note: some Central Asian languages (Uzbek/Tajik) have native speakers in Afghanistan as well (Mobashir, 2021).

evaluation metrics while evaluating the resolved coreferences having split antecedents.

## 2. Related Work

Early works in coreference resolution used constituency trees in Hobbs algorithm (Hobbs, 1978), semantic features (Lappin and Leass, 1994), and syntactic features (Haghighi and Klein, 2009). A mention-ranking architecture of coreference resolution using pretrained word embeddings and neural networks was first proposed by Lee et al. (2017). Each mention is denoted by a span of words (tokens) in the text. Equation 1 is used to calculate a score for a given pair of spans. A dummy antecedent is denoted by the symbol $\epsilon$, and the value of $S(i, j = \epsilon)$ is always taken as zero. The score ($S$) represents the strength of coreference link between the span *i* and *j*. In equation 1, the mention score ($s_m$) and antecedent score ($s_a$) are calculated using the span representations which are obtained with the help of pretrained word embeddings and Bi-LSTM neural network.

$$S(i,j) = s_m(i) + s_m(j) + s_a(i,j) \quad (1)$$

The aim of the model was to learn a conditional probability distribution mentioned in equation 2 where $Y(i)$ represents the set of all the possible mention spans before the span *i* during the discourse.

$$P(y_i) = \frac{e^{S(i,y_i)}}{\sum_{y' \in Y(i)} e^{S(i,y')}} \quad (2)$$

The approach introduced by Lee et al. (2017) has served as a source of inspiration for various studies in the realm of end-to-end coreference resolution. Joshi et al. (2019) observed notable performance improvements when applying a pretrained transformer-based model by Devlin et al. (2019), rather than static word embeddings, for text encoding. Additionally, Joshi et al. (2020) employed a span-based objective to pretrain a transformer-based model, which they used for text encoding and coreference resolution, following the approach by Lee et al. (2017). Meanwhile, Xu and Choi (2020) empirically demonstrated that the Higher-Order Inference (HOI) method proposed by Lee et al. (2018) often has minimal, and at times, even negative, impact on coreference resolution.

In terms of South Asian languages, several early works have explored Hindi coreference resolution, including the work by Dutta et al. (2008), which proposed a modified Hobbs algorithm. Many works have used hand-crafted rules to resolve coreferences in Hindi (Dakwale et al., 2013; Lata et al., 2023), Marathi (Khandale and Mahender, 2019b), and Telugu (Jonnalagadda and Mamidi, 2015). Some works have used Person-Number-Gender (PNG) features to detect mentions and Conditional Random Field (CRF) model to predict coreferential links in Tamil (Akilandeswari and Devi, 2012; Ram and Devi, 2020) and Hindi (Devi et al., 2014). Incorporating PNG features for some South Asian language poses significant challenges due to their distinct inflectional system (Gandhe et al., 2011). Unlike European languages, many South Asian languages inflect verbs according to the actions of the sentence rather than the agents (Shapiro, 2003; Singh and Sarma, 2011). Furthermore, inflectional errors are prevalent in such languages (Sonawane et al., 2020). Therefore, in order to advance multilingual coreference resolution, neural techniques for automated feature extraction are essential. Singh et al. (2020) resolved anaphoras in Hindi using Gated Recurrent Unit (GRU) with static word embeddings. However, no public tool or source code is available for any of the aforementioned works in South Asian languages.

## 3. Dataset

In this research, we incorporated the following manually annotated English coreference resolution datasets: (i) OntoNotes, widely recognized as a benchmark dataset for coreference resolution (Weischedel et al., 2013; Shridhar et al., 2023; Xia and Van Durme, 2021), and (ii) LitBank, which contains longer documents and includes singleton mentions, i.e., mentions that occur only once in the discourse (Recasens et al., 2013; Bamman et al., 2020). It is worth noting that OntoNotes lacks sin-gleton mentions, which led us to utilize the LitBank dataset in our study. Kübler and Zhekova (2011) and Yu et al. (2020) has demonstrated that detecting singleton mentions impacts the performance of a coreference resolution system.

|  |  | #sents | #mentions | #coreference clusters | #split-antecedants | #singletons | #docs |
|---|---|---|---|---|---|---|---|
| TransMuCoRes | Train | 1839883 | 3821540 | 1135906 | 93668 | 350017 | 87946 |
|  | Development | 224911 | 472083 | 148189 | 10505 | 46399 | 10890 |
|  | Test | 255466 | 558093 | 165664 | 12944 | 59279 | 11294 |
| Mujadia et al., 2016 | Train | 2839 | 10512 | 3217 | 287 | 538 | 220 |
|  | Development | 347 | 1306 | 387 | 31 | 58 | 27 |
|  | Test | 347 | 1255 | 399 | 36 | 79 | 28 |

Table 2: Data statistics for TransMuCoRes across 31 South Asian languages, and data statistics of the Mujadia et al. (2016) dataset in Hindi. It can be observed that the ratio of #split-antecedent with #mentions is similar in both the datasets, with percentages of 2.4% and 2.7% for TransMuCoRes and the Mujadia et al. (2016) dataset, respectively.

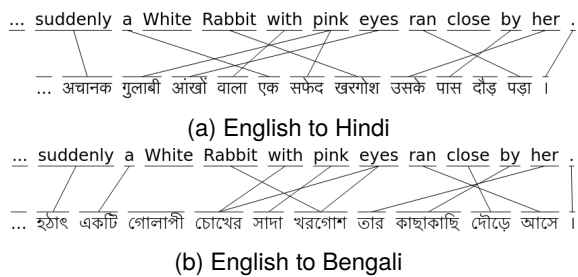

(a) English to Hindi

(b) English to Bengali

Figure 2: Visualizations of word-alignments predicted by the fine-tuned multilingual checkpoint by Dou and Neubig (2021) in high-recall setting. It can be observed that word-order of Hindi and Bengali is different than English.

|  | Mentions | | |
|---|---|---|---|
|  | **Aligned** | **Misaligned** | **Non-Aligned** |
| simalign with multi-lingual BERT (mbert) | 53.7% | **6.1%** | 40.1% |
| simalign with XLM-RoBERTa (xlmr) | 58.5% | 7.1% | 34.3% |
| awesome-align without high recall | 66.7% | 9.4% | 23.8% |
| awesome-align with high recall | **72.5%** | 11.4% | **16.2%** |

Table 3: Alignment statistics from awesome-align (Dou and Neubig, 2021) and simalign (Sabet et al., 2020) on TransMuCoRes shows that high-recall checkpoint of awesome-align method gives most number of aligned mentions.

Figure 1 provides an overview of the pipeline utilized for constructing individual samples within TransMuCoRes. We used *nllb-200-1.3B* model (Team et al., 2022) to translate the English sentences to its target language. Table 1 shows the languages supported by TransMuCoRes. We maintained a sanity-check for the generated translations, considering it a failure if the translation primarily consisted of repeated punctuation. The English sentences whose generated translations failed the sanity check were re-translated using the larger *facebook/nllb-200-3.3B* model. After this we observed that only 111 translations failed the sanity-check out of more than 3 million translations. Nearly 12% sanity-check failures were from Sindhi language. Appendix A contains the language-wise distribution of sanity-check in Table 7.

When translating an English sentence into South Asian languages, the position of mentions within the translated sentence can change due to the free word order characteristics of these languages (Dayal and Mahajan, 2004). To illustrate this, consider an excerpt from a sentence in LitBank: "... suddenly a White Rabbit with pink eyes ran close by her." This sentence is translated into Hindi (... अचानक गुलाबी आंखों वाला एक सफेद खरगोश उसके पास दौड़ पड़ा ।) and Bengali (... হঠাৎ একটি গোলাপী চোখের সাদা খরগোশ তার কাছাকাছি দৌড়ে আসে ।) using the NLLB model. Figure 2 provides a visual representation of the free-word nature of Hindi and Bengali in these translations[3].

In order to generate word-level alignments after the translation step, we used the high-recall multilingual checkpoint from the awesome-align tool (Dou and Neubig, 2021). An aligned mention refers to a continuous span of words in the target language that corresponds to a mention in the source (English) language. If a mention is aligned to a non-continuous span of words in the target language, it is termed a misaligned mention. In cases where a mention lacks alignment with any word in the target language, it is categorized as a non-aligned mention. Misaligned and non-aligned mentions are not marked as mentions in our work. We observed that it gave more aligned mentions as compared to Sabet et al. (2020). Table 3 shows a comparison of the two methods. Figure 2 con-tains Hindi and Bengali translations because we observed that most aligned mentions were seen in these languages. Appendix A contains the alignment statistics for each language in Table 8.

---

[3]Visualizations created by https://vilda.net/s/slowalign/

| | | | |
|---|---|---|---|
| **English** | ... suddenly [a White Rabbit with pink eyes]$_4$ ran close by [her]$_1$ . | **mni_Beng** | ... নোংমদা মচুগী মমি লৈবা ৱাইত রেবিন অমনা মহাক্কী মনাক্তা চংলকখি । |
| **asm_Beng** | ... হঠাতে এটা [ৰঙা চকু থকা বগা কণী]$_4$ তাইৰ ওচৰলৈ দৌৰি আহিল । | **mya_Mymr** | ... [ရုတ်တရက် ပန်းရောင်မျက်လုံးများ ယုန်ဖြူတစ်ကောင်]$_4$ သူမနားကို ပြေးလာပါတယ်။ |
| **awa_Deva** | ... एकाएक उ एक सफेद खरगोश स भरा गुलाबी आँखी क ओकरै लगे [दौड़िके]$_1$ आवा । | **npi_Deva** | ... अचानक गुलाबी आँखा भएको सेतो खरगोश [उनको]$_1$ नजिकै दौड्यो । |
| **ben_Beng** | ... হঠাৎ একটি গোলাপী চোখের সাদা খরগোশ [তার]$_1$ কাছাকাছি দৌড়ে আসে । | **ory_Orya** | ... [ହଠାତ୍ ଏକ ଗୋଲାପୀ ଆଖିର]$_4$ ଶ୍ୱାଇଟ ରାବିଟ୍ ତାଙ୍କ ପାଖରେ ଦୌଡ଼ି ଆସିଲା । |
| **bho_Deva** | ... अचानक उहो [गुलाबी आँखिन क एक सफेद]$_4$ खरगोश ओकरे लगे दौड़त आवा । | **pan_Guru** | ... अचानक [ਇੱਕ ਗੁਲਾਬੀ ਅੱਖਾਂ ਵਾਲਾ ਵ੍ਹਾਈਟ ਰੈਬਿਟ]$_4$ [ਉਸਦੇ]$_1$ ਨੇੜੇ ਭੱਜਿਆ । |
| **bod_Tibt** | ... [གློ་བུར་དུ་མིག་དམར་པོ་དམར་པོ་ཡོད་པའི་རི་བོང་དཀར་པོ་ཞིག་ཐེངས་ཤིག་ཁྱོད་ཀྱི་འགྲམ་དུ་ཡོད་རེད།]$_4$ | **pbt_Arab** | خرګوښ سپین يو ناڅاپه چې کله ... ورغله نږدې سره سترګو ګلابي د |
| **dzo_Tibt** | ... མོང་རྣམས་ (མོང་ཚིགས་པོ་ལས་) མོང་རྣམས་ (མོང་ཚིགས་པོ་ཡོད་པ་) མོང་རྣམས་ | **prs_Arab** | با سفید خرګوش [یک ناګهان که وقتی ... دوید [او]$_1$ نزدیک به صورتی های [چشم]$_4$ |
| **guj_Gujr** | ... અચાનક [એક ગુલાબી આંખોવાળો]$_4$ સફેદ સસલું [તેની]$_1$ નજીક દોડ્યો. | **sat_Beng** | ... ᱥᱟᱦᱮᱵ ᱥᱮᱛᱟᱜ ᱵᱷᱤᱛᱤ ᱠᱟᱱ ᱥᱟᱦᱮᱵ ᱠᱟᱱ ᱵᱷᱤᱛᱤ ᱥᱟᱦᱮᱵ ᱛᱟᱦᱮᱱ ᱮᱱᱟ ᱛᱟᱦᱮᱱ ᱮᱱᱟ । |
| **hin_Deva** | ... अचानक [गुलाबी आंखों वाला एक सफेद खरगोश]$_4$ [उसके]$_1$ पास दौड़ पड़ा । | **sin_Sinh** | ... රෝස [පැහැති ඇස් ඇති]$_4$ සුදු රැහිණෙක් ඇය අසල දිව ගා විට . |
| **hne_Deva** | ... अचानक गुलाबी आंखी वाला [एक सफेद खरगोश]$_4$ ओखर करा दौड़त आईस । | **snd_Arab** | ګلابي کنپيءَ اچو هڪ اوچتو جڏهن ... دوڙندو ويجهو جي هن سان اکين |
| **kan_Knda** | ... ಇದ್ದಕ್ಕಿದ್ದಂತೆ [ಗುಲಾಬಿ ಕಣ್ಣಿನ ಬಿಳಿ ಮೊಲವು]$_4$ [ಅವಳ]$_1$ ಹತ್ತಿರ ಓಡಿತು . | **tam_Taml** | ... ஒரு வெள்ளை முயல் ரோஜா கண்கள்டன் [அவளிடம்]$_1$ நெருங்கி ஓடியது . |
| **kas_Arab** | خرګوش سفيد اکھ اچانک رلان. ... نزدیک تیلٕہ اوس ستی چشمن ګلابي ییمٕہ | **tel_Telu** | ... [ఒక బ్లూ-కళ్ళు గల]$_4$ విట్ కనేక్ట్ [ఆమె]$_1$ ధగ్గరకు వచ్చింది . |
| **mag_Deva** | ... अचानक एगो गुलाबी [आँख वाला सफेद खरगोश]$_4$ ओकरा नगीच दौड़लइ । | **tgk_Cyrl** | ... ки ногаҳон [як харгӯши сафед бо чашмони гулобӣ]$_4$ ба наздикии [ӯ]$_1$ давида омад . |
| **mai_Deva** | ... अचानक [एकटा गुलाबी-आँखि]$_4$ वला गोरगो ओकर लगमे दौड़ि [गेल]$_1$ । | **uig_Arab** | ئويلنسپ ئۇ ، باقتىغۇ ئويلنسپ ... باقتىغۇ ئويلنسپ ئۇ ، |
| **mal_Mlym** | ... പെട്ടെന്ന് ഒരു റോസ് കണ്ണുകളുള്ള വൈറ്റ് കൺകീറ്റ് [അവളുടെ]$_1$ അടുത്ത് ഓടി . | **urd_Arab** | آنکھوں گلابی خرګوش سفيد ایک اچانک جب ... گیا. ساتھ کے [اس]$_1$ کے قریب بھاگ |
| **mar_Deva** | ... अचानक [एक गुलाबी डोळे असलेला पांढरा ससा]$_4$ [तिच्या]$_1$ जवळ धावला . | **uzn_Latn** | ... [unga]$_1$ yaqinlashib ketayotgan qizg'al ko'zli Oq quyon ko'zidan o'tib qoldi . |

Table 4: Coreference data after processing the translated sentences and aligned words. Coreference clusters for "*Rabbit*" and "*Alice*" are highlighted in green and pink color, respectively. The English sentence in this table is an excerpt from a sentence in LitBank (Bamman et al., 2020). Translation errors can be observed in dzo_Tibt where words are repeated. Alignment errors can be observed in ben_Beng where the mention "*a White Rabbit with pink eyes*" is misaligned despite perfect translation. Note: Arabic fonts are to be read left to right due to issues in the LaTeX typefonts.

The proposed TransMuCoRes was constructed using the following three primary components: (i) manually annotated mentions and coreference clusters in the English dataset, (ii) predicted translations, and (iii) aligned mentions between English sentences and translated sentences. For each language, dummy values were assigned in the placeholder of constituency parse trees and speaker information. This was necessitated by the absence of publicly available tools for generating constituency parse trees for South Asian languages and the lack of speaker information in the Litbank and Mujadia et al. (2016) dataset.

Table 4 highlights errors in translation and alignment observed in the TransMuCoRes construction pipeline. Due to imperfect translations and alignments, we opted to train various off-the-shelf coreference resolution models with a manually annotated coreference resolution dataset as well. We utilized the Hindi coreference resolution dataset introduced by Mujadia et al. (2016), because it is the only publicly available manually annotated dataset for coreference resolution in a South Asian language (Hindi). For details on the statistics of TransMuCoRes and the dataset by Mujadia et al. (2016), please refer to Table 2. While we retained the train:dev:test splits for OntoNotes, we had to establish similar splits for LitBank and the dataset by Mujadia et al. (2016). The splits will be released as part of the resources to encourage a standardized evaluation of future models.

## 4. Coreference Resolution Models

In this study, we used the following off-the-shelf coreference resolution models: (i) wl-coref (Dobrovolskii, 2021), and (ii) fast-coref (Toshniwal et al., 2021). We selected these models primarily because they offer fine-tuning scripts for a new CoNLL formatted data. Moreover, wl-coref and fast-coref delivers performances only 3-4% lower than the state-of-the-art model on the OntoNotes benchmark[4]. Unfortunately, we encountered challenges in finding comparable training scripts for state-of-the-art coreference resolution models on the OntoNotes dataset (Liu et al., 2022; Bohnet et al., 2023; Werlen and Henderson, 2022). Due to limitations in available resources, fine-tuning of the LingMess model (Otmazgin et al., 2023) was not pursued in this study. Similarly, other coreference models (Otmazgin et al., 2022; Kirstain et al., 2021), which have not demonstrated superior performance compared to wl-coref and fast-coref on OntoNotes benchmark, were also not subjected to fine-tuning in our investigation. Furthermore, coreference models such as those proposed by Aloraini et al. (2020) and Yu et al. (2020) relied on features extracted from pretrained word embeddings specific to individual languages, rendering the process of fine-tuning for multilingual data less straightforward. To adapt wl-coref and fast-coref for multilingual data, we harnessed the capabilities of multilingual BERT (mbert) by Devlin et al. (2019) and base XLM-RoBERTa (xlmr) by Conneau et al. (2020) as text encoders. This approach allowed for the fine-tuning of a single model with multilingual data.

The fast-coref model is constructed through the utilization of the Longformer encoder (Beltagy et al., 2020) within the longdoc coreference resolution framework (Toshniwal et al., 2020). It was noted that concurrent training employing data augmentation methodologies such as pseudo-singletons enhanced the model's performance across various datasets spanning diverse domains.

---

[4] https://paperswithcode.com/sota/coreference-resolution-on-ontonotes

The wl-coref model relies on a head-finding mechanism via dependency parse trees, which compelled us to fine-tune the model exclusively for languages with publicly available dependency parsers. Therefore, we fine-tuned wl-coref model on Hindi, Tamil, Telugu, Urdu, and Marathi data using the dependency parser of Stanza (Qi et al., 2020) library. The fine-tuned wl-coref model was evaluated in zero-shot manner for the remaining languages.

It is worth noting that the Stanza dependency parsing tool frequently generates parse trees by subdividing words into subword components. For instance, in the tokenization of the sentence, கடுமையான வலிகளும் , திடீர் தலைச்சுற்றலும் , பின்னர் துளைகளிலிருந்து இரத்தப்போக்கு , மற்றும் உடைப்புகளும் இருந்தன . *(There were severe pains, sudden dizziness, then bleeding from the pores, and ruptures.)*, the word கடுமையான *(severe)* is broken into கடுமைய் *(severely)* and ஆன *(became)*. It becomes evident that a mere concatenation of subwords does not yield an equivalent representation to the original token. It is important to underline that the TransMuCoRes dataset exclusively provides word-level annotations. Therefore, obtaining a dependency parse tree at the word level, rather than at the subword level, holds significant relevance. To address this challenge, we harnessed the utility of the awesome-align word alignment tool to establish a mapping between subwords and their corresponding original words.

### 4.1. Evaluation Metrics

In this study, we employ evaluation metrics traditionally utilized for coreference resolution (Joshi et al., 2020; Toshniwal et al., 2021; Paun et al., 2022). MUC functions as a link-based metric, where a lower MUC value indicating a substantial need for links to be added or deleted in the predicted coreference chains to align them closely with the ground-truth coreference chain. Conversely, $B^3$ is a mention-based metric that assesses how effectively the coreference model groups together corefering mentions while keeping non-corefering mentions distinct. CEAFe reflects the degree of overlap between aligned key-response pairs, with a higher value indicating greater alignment between key and response. The CoNLL metric is derived from an unweighted average of MUC, $B^3$, and CEAFe. Furthermore, LEA is a link and entity-based metric, with a higher LEA value indicating accurate resolution of long coreference chains. These metrics have been elaborated in more detail by (Moosavi and Strube, 2016).

|  |  |  |  | Mentions | | | MUC | | | B³ | | | CEAFe | | | LEA | | | CoNLL |
|---|---|---|---|---|---|---|---|---|---|---|---|---|---|---|---|---|---|---|---|
|  |  |  |  | P | R | F1 | P | R | F1 | P | R | F1 | P | R | F1 | P | R | F1 | F1 |
| wl-coref (Dobrovolskii, 2021) | 5 langs | mbert | dev | 62 | **77** | 69 | 57 | **67** | 61 | 42 | **55** | 47 | 37 | 55 | 44 | 36 | 49 | 42 | 51 |
|  |  |  | test | 64 | **79** | 71 | 61 | 70 | 65 | 42 | **56** | 48 | 34 | 53 | 41 | 37 | **51** | 43 | 52 |
|  |  | xlmr | dev | **67** | **77** | **72** | **63** | 67 | 65 | **49** | **55** | **52** | **41** | **59** | **48** | **43** | **50** | **46** | **55** |
|  |  |  | test | **68** | **79** | **73** | **66** | **71** | **69** | **49** | **56** | **52** | 37 | **58** | **45** | **44** | **51** | **48** | **55** |
|  | All langs | mbert | dev | 37 | 68 | 48 | 32 | 56 | 41 | 22 | **46** | 30 | 20 | 45 | 28 | 18 | **40** | 25 | 33 |
|  |  |  | test | 39 | 70 | 50 | 34 | 59 | 44 | 22 | **47** | 30 | 19 | 43 | 26 | 18 | **42** | 26 | 33 |
|  |  | xlmr | dev | 45 | 62 | 52 | 40 | 51 | 45 | 29 | 41 | 34 | **25** | 45 | **32** | 25 | 36 | 29 | 37 |
|  |  |  | test | 46 | 63 | 53 | 42 | 54 | 47 | 29 | 41 | 34 | **23** | 44 | **30** | 25 | 37 | 30 | 37 |
| fast-coref (Toshniwal et al., 2021) | All langs | mbert | dev | 44 | **76** | 56 | 41 | 59 | 48 | 28 | 42 | 34 | 18 | 55 | 27 | 24 | 36 | 29 | 36 |
|  |  |  | test | 46 | 76 | 58 | 44 | 62 | 52 | 29 | 42 | 34 | 17 | 53 | 26 | 25 | 37 | 30 | 37 |
|  |  | xlmr | dev | **48** | **76** | **59** | **46** | **61** | **52** | **33** | **44** | **38** | 21 | **59** | 31 | **29** | 39 | **33** | **41** |
|  |  |  | test | **50** | **77** | **60** | **49** | **64** | **56** | **34** | **44** | **38** | 20 | **58** | 29 | **30** | **40** | **34** | **41** |

Table 5: Performance of fine-tuned fast-coref (Toshniwal et al., 2021) with xlmr encoder is better than zero-shot performance of wl-coref (Dobrovolskii, 2021) on all the languages. However, wl-coref is found to be performing well for the 5 languages on which it is fine-tuned.

| Language | Split | fast-coref (Toshniwal et al., 2021) vs wl-coref (Dobrovolskii, 2021) on fine-tuned xlmr | | | | | | Split | | | | | | |
|---|---|---|---|---|---|---|---|---|---|---|---|---|---|---|
|  |  | Mentions F1 | MUC F1 | B³ F1 | CEAFe F1 | LEA F1 | CoNLL F1 |  | Mentions F1 | MUC F1 | B³ F1 | CEAFe F1 | LEA F1 | CoNLL F1 |
| asm_Beng | DEVELOPMENT | 45 vs 46 | 38 vs 35 | 25 vs 24 | 22 vs 26 | 20 vs 18 | 28 vs 28 | TEST | 47 vs 46 | 41 vs 37 | 25 vs 23 | 21 vs 24 | 20 vs 18 | 29 vs 28 |
| awa_Deva |  | 50 vs 42 | 41 vs 32 | 29 vs 25 | 27 vs 28 | 23 vs 19 | 33 vs 28 |  | 51 vs 43 | 45 vs 33 | 28 vs 23 | 25 vs 26 | 23 vs 18 | 33 vs 27 |
| ben_Beng |  | 74 vs 73 | 67 vs 66 | 51 vs 52 | 45 vs 48 | 46 vs 47 | 55 vs 55 |  | 76 vs 75 | 71 vs 70 | 52 vs 53 | 43 vs 46 | 48 vs 49 | 56 vs 56 |
| bho_Deva |  | 52 vs 41 | 46 vs 33 | 31 vs 24 | 27 vs 26 | 26 vs 19 | 35 vs 27 |  | 54 vs 43 | 49 vs 35 | 31 vs 24 | 25 vs 24 | 26 vs 19 | 35 vs 28 |
| bod_Tibt |  | 63 vs 6 | 44 vs 3 | 17 vs 2 | 10 vs 3 | 12 vs 2 | 24 vs 3 |  | 64 vs 6 | 48 vs 3 | 18 vs 3 | 10 vs 4 | 14 vs 2 | 25 vs 3 |
| dzo_Tibt |  | 18 vs 4 | 15 vs 3 | 9 vs 2 | 5 vs 2 | 6 vs 1 | 10 vs 2 |  | 19 vs 5 | 16 vs 4 | 10 vs 2 | 6 vs 3 | 7 vs 2 | 10 vs 3 |
| guj_Gujr |  | 74 vs 73 | 66 vs 65 | 50 vs 51 | 45 vs 48 | 45 vs 46 | 54 vs 55 |  | 75 vs 74 | 69 vs 68 | 50 vs 51 | 42 vs 45 | 45 vs 47 | 54 vs 55 |
| hin_Deva |  | 75 vs 74 | 68 vs 68 | 52 vs 54 | 46 vs 50 | 47 vs 49 | 55 vs 58 |  | 76 vs 76 | 71 vs 72 | 52 vs 55 | 43 vs 48 | 48 vs 51 | 55 vs 58 |
| hne_Deva |  | 48 vs 39 | 41 vs 29 | 28 vs 21 | 27 vs 24 | 23 vs 16 | 32 vs 25 |  | 48 vs 40 | 43 vs 31 | 28 vs 21 | 25 vs 22 | 23 vs 16 | 32 vs 25 |
| kan_Knda |  | 73 vs 71 | 65 vs 63 | 50 vs 50 | 45 vs 48 | 44 vs 45 | 53 vs 54 |  | 74 vs 73 | 68 vs 67 | 50 vs 51 | 42 vs 46 | 46 vs 46 | 53 vs 55 |
| kas_Arab |  | 33 vs 13 | 27 vs 8 | 17 vs 7 | 16 vs 10 | 13 vs 5 | 20 vs 8 |  | 34 vs 14 | 30 vs 9 | 17 vs 6 | 15 vs 9 | 13 vs 4 | 20 vs 8 |
| mag_Deva |  | 54 vs 47 | 47 vs 38 | 32 vs 28 | 28 vs 28 | 27 vs 22 | 36 vs 31 |  | 56 vs 50 | 51 vs 41 | 32 vs 27 | 26 vs 27 | 28 vs 22 | 36 vs 32 |
| mai_Deva |  | 48 vs 38 | 42 vs 32 | 28 vs 22 | 25 vs 22 | 24 vs 18 | 32 vs 25 |  | 50 vs 39 | 45 vs 34 | 29 vs 22 | 23 vs 21 | 25 vs 18 | 32 vs 25 |
| mal_Mlym |  | 66 vs 65 | 58 vs 56 | 43 vs 44 | 38 vs 44 | 37 vs 37 | 46 vs 48 |  | 68 vs 67 | 62 vs 60 | 44 vs 45 | 36 vs 42 | 39 vs 39 | 47 vs 49 |
| mar_Deva |  | 72 vs 71 | 65 vs 64 | 49 vs 51 | 44 vs 47 | 44 vs 46 | 52 vs 54 |  | 74 vs 72 | 68 vs 68 | 50 vs 52 | 40 vs 44 | 45 vs 47 | 53 vs 54 |
| mni_Beng |  | 30 vs 9 | 24 vs 4 | 14 vs 3 | 11 vs 5 | 9 vs 1 | 16 vs 4 |  | 33 vs 10 | 27 vs 6 | 14 vs 3 | 10 vs 5 | 10 vs 1 | 17 vs 4 |
| mya_Mymr |  | 61 vs 53 | 52 vs 42 | 37 vs 32 | 34 vs 35 | 31 vs 26 | 41 vs 36 |  | 63 vs 54 | 55 vs 46 | 36 vs 32 | 31 vs 33 | 31 vs 26 | 41 vs 37 |
| npi_Deva |  | 74 vs 73 | 68 vs 66 | 52 vs 53 | 46 vs 49 | 47 vs 47 | 55 vs 56 |  | 76 vs 75 | 71 vs 70 | 53 vs 54 | 44 vs 47 | 49 vs 49 | 56 vs 57 |
| ory_Orya |  | 22 vs 22 | 19 vs 13 | 10 vs 9 | 8 vs 12 | 7 vs 4 | 12 vs 11 |  | 24 vs 23 | 22 vs 15 | 10 vs 8 | 8 vs 12 | 7 vs 4 | 13 vs 12 |
| pan_Guru |  | 71 vs 70 | 64 vs 63 | 48 vs 49 | 43 vs 47 | 43 vs 44 | 51 vs 53 |  | 73 vs 72 | 68 vs 67 | 48 vs 50 | 40 vs 44 | 44 vs 45 | 52 vs 53 |
| pbt_Arab |  | 33 vs 29 | 31 vs 21 | 18 vs 14 | 15 vs 17 | 15 vs 9 | 21 vs 17 |  | 36 vs 31 | 34 vs 23 | 19 vs 13 | 14 vs 17 | 16 vs 9 | 22 vs 18 |
| prs_Arab |  | 69 vs 64 | 63 vs 56 | 47 vs 43 | 40 vs 42 | 42 vs 37 | 50 vs 47 |  | 70 vs 66 | 66 vs 60 | 46 vs 44 | 37 vs 39 | 42 vs 39 | 50 vs 48 |
| sat_Beng |  | 11 vs 4 | 10 vs 2 | 5 vs 2 | 3 vs 2 | 4 vs 1 | 6 vs 2 |  | 13 vs 5 | 12 vs 4 | 6 vs 2 | 3 vs 2 | 4 vs 1 | 7 vs 3 |
| sin_Sinh |  | 22 vs 22 | 18 vs 13 | 10 vs 9 | 8 vs 13 | 7 vs 5 | 12 vs 11 |  | 24 vs 23 | 22 vs 15 | 11 vs 9 | 8 vs 12 | 8 vs 5 | 14 vs 12 |
| snd_Arab |  | 30 vs 31 | 26 vs 22 | 15 vs 14 | 13 vs 18 | 12 vs 9 | 18 vs 18 |  | 31 vs 32 | 28 vs 24 | 15 vs 14 | 12 vs 18 | 12 vs 10 | 18 vs 18 |
| tam_Taml |  | 71 vs 70 | 63 vs 63 | 47 vs 50 | 43 vs 49 | 42 vs 44 | 51 vs 54 |  | 72 vs 72 | 66 vs 66 | 48 vs 50 | 40 vs 45 | 43 vs 45 | 51 vs 54 |
| tel_Telu |  | 71 vs 71 | 64 vs 64 | 48 vs 51 | 43 vs 48 | 43 vs 46 | 52 vs 54 |  | 73 vs 73 | 68 vs 68 | 50 vs 52 | 39 vs 45 | 45 vs 48 | 52 vs 55 |
| tgk_Cyrl |  | 61 vs 16 | 52 vs 11 | 37 vs 8 | 33 vs 11 | 31 vs 6 | 41 vs 10 |  | 62 vs 15 | 55 vs 11 | 37 vs 7 | 30 vs 10 | 32 vs 5 | 41 vs 10 |
| uig_Arab |  | 21 vs 25 | 15 vs 13 | 9 vs 10 | 9 vs 15 | 6 vs 5 | 11 vs 13 |  | 21 vs 26 | 17 vs 16 | 8 vs 10 | 8 vs 14 | 6 vs 6 | 11 vs 13 |
| urd_Arab |  | 70 vs 71 | 63 vs 64 | 47 vs 50 | 41 vs 47 | 42 vs 45 | 51 vs 54 |  | 72 vs 72 | 67 vs 67 | 47 vs 50 | 39 vs 44 | 43 vs 45 | 51 vs 54 |
| uzn_Latn |  | 63 vs 60 | 54 vs 50 | 40 vs 39 | 38 vs 40 | 34 vs 33 | 44 vs 43 |  | 64 vs 60 | 57 vs 53 | 39 vs 38 | 35 vs 37 | 34 vs 33 | 43 vs 43 |
| Mujadia et al. (2016) |  | 50 vs 78 | 45 vs 74 | 35 vs 66 | 33 vs 62 | 31 vs 62 | 38 vs 67 |  | 54 vs 79 | 51 vs 76 | 38 vs 68 | 31 vs 60 | 34 vs 64 | 40 vs 68 |
| Overall |  | 59 vs 52 | 52 vs 45 | 38 vs 34 | 31 vs 32 | 33 vs 29 | 41 vs 37 |  | 60 vs 53 | 56 vs 47 | 38 vs 34 | 29 vs 30 | 34 vs 30 | 41 vs 37 |

Table 6: The wl-coref (Dobrovolskii, 2021) method performs better than fast-coref (Toshniwal et al., 2021) for the languages on which it was fine-tuned (Hindi, Tamil, Telugu, Urdu, and Marathi).

## 5. Results

In this study, we utilized evaluation scripts developed by Paun et al. (2022). Table 5 illustrates the strong performance of the wl-coref method when evaluating languages it was fine-tuned on. This is confirmed from the findings in Table 6 that wl-coref performs better than fast-coref only on those languages on which it was fine-tuned. Notably, wl-coref achieves the highest performance on our test split of the golden set, with scores of 68 for LEA F1 and 64 for CoNLL F1. Our work is the first to release coreference resolution tools for Hindi and report their performance on the golden set by Mujadia et al. (2016). We also observed that the performance of both the models improves when singletons are ignored during the evaluation. Table 10 and Table 11 in Appendix A illustrates the same. In order to reconstruct the proposed TransMuCoRes, 14 GB of GPU memory is required, and the pro-cess may extend up to three months in duration on a single GPU. Similarly, the fine-tuning of coreference resolution models mandates a minimum of 30 GB of GPU memory and may span up to eight hours of computational runtime. Section A.1 in Appendix A contains details about the compute resources needed for this work.

We have observed that one of the CoNLL metrics, BCUB (BAGGA, 1998), may fail to generate meaningful scores on instances containing splitted antecedants in the key coreference chains. For instance consider the following paragraph:

> **Thatcher**$_a$ grew up in Lincolnshire whereas **Gandhi**$_b$ was raised in Allahabad. **Both**$_c$ become powerful figures. **They**$_d$ locked horns in 1983. The world watched as **the Iron Lady of India**$_e$ stood against **the Iron Lady of UK**$_f$.

If a system response (predictions) is identical to the key (ground-truth), then the BCUB recall score would be 1.25. Furthermore, the LEA metric, introduced as an alternative to CoNLL metrics by Moosavi and Strube (2016) also yields imperfect scores for the above example. The formulation of LEA metric gives a score of 1.16 for recall, precision, and F1 when the ground-truth is used as the key and response. This underscores the necessity for an evaluation metric capable of effectively handling split antecedents in coreference resolution. Furthermore, the presence of split antecedents not only highlights a significant deficiency in current coreference resolution evaluation metrics but also poses challenges in the training of existing coreference resolution models. This challenge stems from architectural limitations in these models, which permit only one antecedent per mention, whereas split antecedents refer to multiple antecedents earlier in the discourse.

## 6. Conclusion

Numerous research initiatives have tackled the automated resolution of coreferences in South Asian languages. However, a notable absence of publicly available resources and models still persists in this domain. This study addresses this gap by introducing TransMuCoRes, a Translated dataset designed for Multilingual Coreference Resolution. We also release checkpoints of two off-the-shelf methods fine-tuned on TransMyCoRes and the golden set of Hindi. Our observations indicate that fine-tuning the wl-coref method is feasible for specific South Asian languages that have an available dependency parser. Notably, it outperforms fast-coref in the languages on which it was fine-tuned.

### 6.1. Limitations and Future Work

This work encounters a significant constraint concerning the potential for transferring bias during the translation of the English dataset into various target languages (Cao and Daumé III, 2020). Furthermore, substantial computational resources are essential to replicate this study. The assessment of coreference resolution models for languages other than Hindi posed a significant challenge attributable to the scarcity of gold annotated resources. We acknowledge the constrained scope of the dataset available for evaluating languages other than Hindi.

The observations presented here are contingent on the specific data splits chosen from LitBank and the dataset by Mujadia et al. (2016). To validate these findings, cross-validation experiments are required. In the future, we aim to expand the dataset to encompass additional low-resource languages supported by NLLB models. Additionally, there is a pressing need for the development of a new evaluation metric capable of accurately assessing coreference clusters that involve split antecedents.

There are some future directions that needs to be explored to improve the performance of wl-coref model for other languages. Data preprocessing for training the wl-coref model requires dependency parsing, primarily to identify the syntactic headword of each mention. In cases where a dependency parsing tool is not accessible for specific languages, alternative approaches can be investigated to identify the head-word of a mention[5]. Accurate identification of head-words would facilitate fine-tuning the wl-coref model across diverse languages, thereby enhancing its efficacy.

## 7. Acknowledgements

Ritwik Mishra expresses gratitude to the University Grants Commission (UGC) of India for providing partial support through the UGC Senior Research Fellowship (SRF) program. Rajiv Ratn Shah acknowledges the partial assistance received from the Infosys Center for AI, the Center of Design and New Media, and the Center of Excellence in Healthcare at IIIT Delhi.

## 8. Bibliographical References

---
[5] https://github.com/vdobrovolskii/wl-coref/issues/12

## 9. Language Resource References

| Language | Translation Sanity-check | |
|---|---|---|
| | Passed | Failed |
| asm_Beng | 102828 | 3 |
| awa_Deva | 102828 | 3 |
| ben_Beng | 102825 | 6 |
| bho_Deva | 102829 | 2 |
| bod_Tibt | 102828 | 3 |
| dzo_Tibt | 102827 | 4 |
| guj_Gujr | 102828 | 3 |
| hin_Deva | 102827 | 4 |
| hne_Deva | 102829 | 2 |
| kan_Knda | 102828 | 3 |
| kas_Arab | 102831 | 0 |
| mag_Deva | 102827 | 4 |
| mai_Deva | 102828 | 3 |
| mal_Mlym | 102826 | 5 |
| mar_Deva | 102826 | 5 |
| mni_Beng | 102829 | 2 |
| mya_Mymr | 102827 | 4 |
| npi_Deva | 102828 | 3 |
| ory_Orya | 102829 | 2 |
| pan_Guru | 102829 | 2 |
| pbt_Arab | 102830 | 1 |
| prs_Arab | 102827 | 4 |
| sat_Beng | 102828 | 3 |
| sin_Sinh | 102829 | 2 |
| snd_Arab | 102817 | 14 |
| tam_Taml | 102827 | 4 |
| tel_Telu | 102823 | 8 |
| tgk_Cyrl | 102826 | 5 |
| uig_Arab | 102829 | 2 |
| urd_Arab | 102828 | 3 |
| uzn_Latn | 102829 | 2 |
| **Total** | **3187650** | **111** |

Table 7: Number of translated sentences that passed/failed the sanity-check. The sanity-check was composed of identifying whether the translated sentence is a repetitive sequence of punctuation's or not.

## A. Appendix

### A.1. Compute Resources Needed

The GPU memory footprint of awesome-align model, *facebook/nllb-200-1.3B* model, and *facebook/nllb-200-3.3B* model is 2GB, 7GB, and 14GB, respectively. On average, the translation model, and alignment model takes 3 seconds, and 30 milliseconds, respectively for each sentence. Therefore, the estimated time to re-construct TransMuCoRes is nearly 3 months on a single GPU. During fine-tuning phase, the memory footprint of wl-coref model, and fast-coref model is 30GB, and 8GB, respectively. Whereas, during inference phase, it is 5GB and 1 GB, respectively. The wl-coref model takes 45 mins per epoch whereas fast-coref runs for 100K steps in 6 hours.

| Language | Mentions (223583) | | |
|---|---|---|---|
| | Aligned | Misaligned | Non-aligned |
| asm_Beng | 66.1% | 15.3% | 18.6% |
| awa_Deva | 71.3% | 15% | 13.7% |
| ben_Beng | 87.4% | 6.7% | 5.8% |
| bho_Deva | 72.6% | 14.5% | 12.9% |
| bod_Tibt | 69.7% | 4.5% | 25.7% |
| dzo_Tibt | 62.8% | 7.9% | 29.3% |
| guj_Gujr | 86.7% | 7.1% | 6.2% |
| hin_Deva | 87% | 7.8% | 5.1% |
| hne_Deva | 70.8% | 14.6% | 14.6% |
| kan_Knda | 86.3% | 6.6% | 7.1% |
| kas_Arab | 59.8% | 18.7% | 21.4% |
| mag_Deva | 73.6% | 13% | 13.4% |
| mai_Deva | 69.7% | 13.8% | 16.4% |
| mal_Mlym | 78.9% | 8.6% | 12.5% |
| mar_Deva | 84.5% | 7.1% | 8.5% |
| mni_Beng | 57% | 19.7% | 23.3% |
| mya_Mymr | 79.8% | 7.7% | 12.4% |
| npi_Deva | 86.1% | 6.8% | 7.1% |
| ory_Orya | 46.1% | 10% | 43.9% |
| pan_Guru | 85.2% | 9.4% | 5.4% |
| pbt_Arab | 58% | 22.4% | 19.6% |
| prs_Arab | 82.9% | 8.4% | 8.8% |
| sat_Beng | 52.3% | 12.7% | 34.9% |
| sin_Sinh | 48% | 10.1% | 41.9% |
| snd_Arab | 56.7% | 22.3% | 21.1% |
| tam_Taml | 83.5% | 7.4% | 9.1% |
| tel_Telu | 84.7% | 7.6% | 7.6% |
| tgk_Cyrl | 80.9% | 8.9% | 10.2% |
| uig_Arab | 57.2% | 17.9% | 24.9% |
| urd_Arab | 83.4% | 10.9% | 5.7% |
| uzn_Latn | 77.4% | 8.4% | 14.1% |
| **Total** | **72.5%** | **11.4%** | **16.2%** |

Table 8: Performance of word-alignment tool on various languages. A mention is a continuous span of words. It is considered to be aligned if all the words of the mention are aligned to a continuous span of words in the target language. If it is aligned to a discontinuous span of words in the target language then it is called misaligned. And if a mention is not aligned to any word in the target language then it is considered as non-aligned.

| Language | #sents | #mentions | #coreference clusters | #split antecedants | #singletons | #docs |
|---|---|---|---|---|---|---|
| asm_Beng | (58706, 7174, 8194) | (113275, 13882, 16387) | (35734, 4671, 5182) | (2559, 337, 387) | (12986, 1731, 2089) | (2835, 352, 365) |
| awa_Deva | (59296, 7257, 8212) | (123270, 15298, 18003) | (36869, 4791, 5333) | (1627, 176, 200) | (11965, 1542, 1958) | (2840, 352, 366) |
| ben_Beng | (63812, 7830, 8862) | (151825, 18931, 22395) | (39523, 5171, 5834) | (1116, 115, 123) | (8475, 1136, 1643) | (2843, 352, 366) |
| bho_Deva | (59942, 7378, 8344) | (125719, 15709, 18574) | (36989, 4897, 5436) | (1317, 140, 169) | (11967, 1613, 1972) | (2837, 352, 363) |
| bod_Tibt | (58372, 7157, 8046) | (84334, 10549, 12085) | (36918, 4791, 5414) | (22268, 2491, 3085) | (12046, 1603, 2104) | (2840, 351, 363) |
| dzo_Tibt | (56040, 6837, 7684) | (95515, 11736, 13648) | (35182, 4551, 5059) | (11767, 1276, 1584) | (13404, 1795, 2183) | (2827, 351, 364) |
| guj_Gujr | (63687, 7805, 8835) | (150276, 18687, 22182) | (39279, 5125, 5788) | (1135, 112, 130) | (8744, 1142, 1623) | (2843, 352, 366) |
| hin_Deva | (63841, 7820, 8883) | (151922, 18982, 22373) | (39129, 5157, 5775) | (624, 58, 56) | (8862, 1198, 1689) | (2843, 352, 366) |
| hne_Deva | (59393, 7304, 8270) | (122714, 15269, 17977) | (36854, 4856, 5375) | (1378, 146, 214) | (12232, 1621, 1983) | (2843, 352, 364) |
| kan_Knda | (63815, 7822, 8897) | (149004, 18501, 22007) | (39588, 5172, 5862) | (1645, 171, 215) | (8515, 1143, 1653) | (2843, 352, 365) |
| kas_Arab | (55735, 6733, 7765) | (102470, 12520, 14932) | (34261, 4433, 4956) | (2476, 270, 321) | (14038, 1806, 2150) | (2834, 350, 363) |
| mag_Deva | (60451, 7403, 8409) | (127402, 15940, 18708) | (37356, 4922, 5470) | (1420, 146, 179) | (11762, 1578, 2001) | (2841, 352, 365) |
| mai_Deva | (59284, 7260, 8243) | (120323, 14820, 17608) | (36704, 4760, 5341) | (1927, 202, 280) | (12422, 1588, 2034) | (2838, 351, 365) |
| mal_Mlym | (62141, 7633, 8622) | (133610, 16648, 19613) | (38726, 5086, 5678) | (3451, 360, 459) | (9749, 1355, 1786) | (2842, 352, 364) |
| mar_Deva | (63279, 7732, 8783) | (145616, 18050, 21406) | (39152, 5103, 5748) | (1557, 165, 184) | (9199, 1225, 1710) | (2843, 352, 365) |
| mni_Beng | (54556, 6722, 7642) | (96202, 11657, 13953) | (32757, 4266, 4714) | (3230, 383, 451) | (13896, 1861, 2143) | (2829, 351, 362) |
| mya_Mymr | (62366, 7633, 8659) | (131466, 16418, 19073) | (38975, 5088, 5756) | (6403, 707, 935) | (9412, 1208, 1802) | (2844, 351, 366) |
| npi_Deva | (63471, 7784, 8817) | (149191, 18586, 22003) | (39359, 5141, 5817) | (1294, 134, 151) | (8819, 1165, 1671) | (2841, 352, 365) |
| ory_Orya | (47958, 5775, 6626) | (78750, 9367, 11132) | (29454, 3844, 4162) | (2720, 315, 443) | (14003, 1905, 2079) | (2822, 348, 362) |
| pan_Guru | (63448, 7776, 8842) | (148484, 18395, 21907) | (38776, 5035, 5705) | (701, 85, 83) | (9575, 1200, 1733) | (2839, 352, 366) |
| pbt_Arab | (54706, 6582, 7599) | (100295, 12038, 14467) | (32813, 4211, 4687) | (1772, 225, 253) | (13898, 1800, 2125) | (2826, 350, 363) |
| prs_Arab | (62579, 7650, 8722) | (144778, 17816, 21260) | (39040, 5113, 5729) | (1163, 138, 203) | (9825, 1373, 1807) | (2844, 352, 365) |
| sat_Beng | (50952, 6129, 6949) | (89870, 10704, 12679) | (31207, 4049, 4439) | (2676, 324, 405) | (13883, 1835, 2108) | (2827, 349, 361) |
| sin_Sinh | (48903, 5930, 6743) | (81864, 9910, 11615) | (30266, 3947, 4259) | (2866, 364, 439) | (13875, 1867, 2056) | (2818, 350, 362) |
| snd_Arab | (54718, 6673, 7596) | (98325, 11871, 14214) | (32806, 4237, 4691) | (2141, 289, 350) | (14142, 1825, 2153) | (2822, 350, 363) |
| tam_Taml | (63313, 7765, 8798) | (143782, 17914, 21120) | (39279, 5127, 5795) | (1774, 162, 223) | (9165, 1197, 1711) | (2842, 352, 365) |
| tel_Telu | (63707, 7809, 8868) | (146249, 18198, 21652) | (39306, 5150, 5820) | (1738, 176, 201) | (9062, 1230, 1749) | (2844, 352, 366) |
| tgk_Cyrl | (62447, 7650, 8670) | (139803, 17348, 20413) | (38981, 5064, 5701) | (2052, 241, 286) | (9693, 1301, 1742) | (2843, 352, 365) |
| uig_Arab | (54313, 6611, 7563) | (96106, 11645, 13899) | (33532, 4337, 4843) | (3922, 457, 519) | (14085, 1833, 2187) | (2831, 350, 362) |
| urd_Arab | (63010, 7743, 8759) | (146133, 18133, 21421) | (38494, 5053, 5617) | (616, 68, 64) | (10185, 1364, 1826) | (2840, 352, 365) |
| uzn_Latn | (61642, 7534, 8564) | (132967, 16561, 19387) | (38597, 5041, 5678) | (2333, 272, 352) | (10133, 1359, 1809) | (2842, 352, 366) |
| **Total** | (1839883, 224911, 255466) | (3821540, 472083, 558093) | (1135906, 148189, 165664) | (93668, 10505, 12944) | (350017, 46399, 59279) | (87946, 10890, 11294) |

Table 9: Data statistics of TransMuCoRes for each language. The numbers written inside round brackets represents the ( train , development , test ) splits.

| | | | | Mentions | | | MUC | | | $B^3$ | | | CEAFe | | | LEA | | | CoNLL |
|---|---|---|---|---|---|---|---|---|---|---|---|---|---|---|---|---|---|---|---|
| | | | | P | R | F1 | P | R | F1 | P | R | F1 | P | R | F1 | P | R | F1 | F1 |
| wl-coref (Dobrovolskii, 2021) | 5 langs | mbert | dev | 65 | **75** | 70 | 57 | 67 | 61 | 43 | 54 | 48 | 46 | 52 | 49 | 39 | 49 | 44 | 53 |
| | | | test | 68 | **77** | 72 | 61 | 70 | 65 | 44 | **55** | 49 | 46 | 50 | 48 | 41 | **51** | 45 | 54 |
| | | xlmr | dev | **70** | 75 | 72 | 63 | 68 | 65 | 50 | 55 | 52 | 51 | 55 | 53 | 46 | 50 | 48 | **57** |
| | | | test | **72** | 77 | 75 | 66 | 71 | 69 | 52 | 55 | 53 | 50 | 54 | 52 | 48 | 51 | 50 | 58 |
| | All langs | mbert | dev | 40 | 66 | 50 | 32 | 56 | 41 | 23 | **46** | 31 | 28 | 42 | 33 | 20 | **40** | 27 | 35 |
| | | | test | 42 | 68 | 52 | 34 | 59 | 44 | 23 | **47** | 31 | 27 | 40 | 33 | 21 | **42** | 28 | 36 |
| | | xlmr | dev | 48 | 60 | 53 | 40 | 51 | 45 | 31 | 40 | 35 | **34** | 42 | 37 | 27 | 36 | 31 | 39 |
| | | | test | 50 | 62 | 55 | 42 | 54 | 47 | 31 | 41 | 35 | **33** | 41 | 37 | 28 | 37 | 32 | 40 |
| fast-coref (Toshniwal et al., 2021) | All langs | mbert | dev | 47 | **73** | 57 | 41 | 59 | 48 | 30 | 41 | 35 | 25 | 52 | 34 | 26 | 36 | 30 | 39 |
| | | | test | 50 | 74 | 60 | 44 | 62 | 52 | 31 | 42 | 35 | 25 | 51 | 33 | 27 | 37 | 32 | 40 |
| | | xlmr | dev | **51** | 73 | 60 | 46 | 61 | 52 | 35 | 44 | **39** | 29 | **56** | 38 | 32 | 39 | **35** | 43 |
| | | | test | **54** | 75 | 63 | 49 | 64 | 56 | 36 | 44 | **40** | 29 | 55 | 38 | 33 | 40 | **36** | 44 |

Table 10: Performance of wl-coref (Dobrovolskii, 2021) and fast-coref (Toshniwal et al., 2021) while ignoring the singeltons during the evaluation phase. It can be seen that the performance improves for both the models. Indicating that both the models struggles in capturing the singletons.

| Language | | fast-coref (Toshniwal et al., 2021) vs wl-coref (Dobrovolskii, 2021) on fine-tuned xlmr | | | | | | | | | | | |
|---|---|---|---|---|---|---|---|---|---|---|---|---|---|
| | Split | Mentions F1 | MUC F1 | B³ F1 | CEAFe F1 | LEA F1 | CoNLL F1 | Split | Mentions F1 | MUC F1 | B³ F1 | CEAFe F1 | LEA F1 | CoNLL F1 |
| asm_Beng | DEVELOPMENT | 47 vs 46 | 38 vs 35 | 26 vs 24 | 28 vs 29 | 22 vs 19 | 31 vs 29 | TEST | 49 vs 47 | 41 vs 37 | 26 vs 23 | 28 vs 29 | 22 vs 19 | 32 vs 30 |
| awa_Deva | | 51 vs 44 | 42 vs 32 | 29 vs 25 | 33 vs 32 | 25 vs 20 | 35 vs 30 | | 53 vs 44 | 45 vs 33 | 29 vs 23 | 33 vs 31 | 25 vs 19 | 35 vs 29 |
| ben_Beng | | 75 vs 74 | 67 vs 66 | 52 vs 53 | 51 vs 52 | 48 vs 48 | 57 vs 57 | | 78 vs 76 | 71 vs 70 | 53 vs 54 | 51 vs 52 | 50 vs 51 | 59 vs 59 |
| bho_Deva | | 54 vs 42 | 46 vs 33 | 32 vs 24 | 34 vs 29 | 28 vs 20 | 37 vs 29 | | 56 vs 45 | 49 vs 35 | 32 vs 24 | 33 vs 29 | 28 vs 20 | 38 vs 29 |
| bod_Tibt | | 62 vs 7 | 46 vs 3 | 17 vs 2 | 13 vs 5 | 13 vs 2 | 26 vs 3 | | 63 vs 7 | 51 vs 3 | 19 vs 3 | 14 vs 5 | 16 vs 2 | 28 vs 4 |
| dzo_Tibt | | 20 vs 4 | 15 vs 3 | 9 vs 2 | 7 vs 4 | 7 vs 1 | 10 vs 3 | | 21 vs 6 | 16 vs 4 | 10 vs 3 | 9 vs 5 | 8 vs 2 | 12 vs 4 |
| guj_Gujr | | 74 vs 74 | 66 vs 65 | 50 vs 52 | 50 vs 52 | 46 vs 48 | 55 vs 57 | | 76 vs 76 | 69 vs 68 | 51 vs 52 | 49 vs 52 | 47 vs 48 | 56 vs 58 |
| hin_Deva | | 75 vs 75 | 68 vs 68 | 52 vs 55 | 51 vs 55 | 48 vs 51 | 57 vs 59 | | 77 vs 78 | 71 vs 72 | 53 vs 56 | 51 vs 55 | 50 vs 53 | 58 vs 61 |
| hne_Deva | | 49 vs 40 | 41 vs 29 | 29 vs 21 | 34 vs 28 | 25 vs 17 | 34 vs 26 | | 50 vs 42 | 43 vs 31 | 28 vs 22 | 33 vs 28 | 25 vs 18 | 35 vs 27 |
| kan_Knda | | 73 vs 72 | 65 vs 63 | 50 vs 51 | 50 vs 52 | 46 vs 46 | 55 vs 55 | | 76 vs 74 | 68 vs 67 | 51 vs 52 | 49 vs 52 | 47 vs 48 | 56 vs 57 |
| kas_Arab | | 35 vs 14 | 27 vs 8 | 18 vs 7 | 22 vs 13 | 14 vs 6 | 22 vs 9 | | 37 vs 15 | 30 vs 9 | 18 vs 7 | 21 vs 13 | 15 vs 5 | 23 vs 9 |
| mag_Deva | | 55 vs 48 | 47 vs 38 | 33 vs 28 | 34 vs 33 | 29 vs 23 | 38 vs 33 | | 58 vs 51 | 51 vs 41 | 33 vs 28 | 34 vs 32 | 30 vs 24 | 39 vs 34 |
| mai_Deva | | 50 vs 40 | 42 vs 32 | 29 vs 23 | 32 vs 27 | 25 vs 19 | 34 vs 27 | | 53 vs 41 | 45 vs 34 | 30 vs 23 | 31 vs 26 | 26 vs 19 | 36 vs 27 |
| mal_Mlym | | 67 vs 66 | 58 vs 56 | 43 vs 44 | 44 vs 48 | 39 vs 39 | 48 vs 49 | | 70 vs 68 | 62 vs 60 | 45 vs 46 | 44 vs 48 | 40 vs 41 | 50 vs 51 |
| mar_Deva | | 73 vs 72 | 65 vs 64 | 50 vs 52 | 49 vs 52 | 45 vs 48 | 54 vs 56 | | 75 vs 74 | 68 vs 68 | 51 vs 53 | 48 vs 51 | 47 vs 49 | 56 vs 57 |
| mni_Beng | | 32 vs 10 | 24 vs 4 | 14 vs 3 | 16 vs 7 | 10 vs 2 | 18 vs 5 | | 35 vs 11 | 28 vs 6 | 14 vs 3 | 15 vs 6 | 11 vs 2 | 19 vs 5 |
| mya_Mymr | | 62 vs 54 | 52 vs 42 | 37 vs 33 | 39 vs 39 | 32 vs 28 | 43 vs 38 | | 64 vs 56 | 55 vs 46 | 37 vs 32 | 38 vs 39 | 32 vs 28 | 43 vs 39 |
| npi_Deva | | 75 vs 74 | 68 vs 66 | 53 vs 53 | 52 vs 53 | 49 vs 49 | 57 vs 57 | | 78 vs 76 | 72 vs 70 | 54 vs 55 | 52 vs 54 | 51 vs 51 | 59 vs 60 |
| ory_Orya | | 25 vs 22 | 19 vs 13 | 11 vs 8 | 13 vs 13 | 9 vs 5 | 14 vs 11 | | 28 vs 23 | 22 vs 15 | 11 vs 8 | 13 vs 13 | 9 vs 5 | 15 vs 12 |
| pan_Guru | | 72 vs 71 | 64 vs 63 | 48 vs 50 | 48 vs 51 | 44 vs 45 | 53 vs 54 | | 74 vs 73 | 68 vs 67 | 49 vs 51 | 48 vs 50 | 46 vs 47 | 55 vs 56 |
| pbt_Arab | | 36 vs 29 | 31 vs 21 | 20 vs 13 | 22 vs 19 | 17 vs 10 | 24 vs 18 | | 39 vs 31 | 34 vs 23 | 20 vs 13 | 22 vs 20 | 18 vs 10 | 25 vs 19 |
| prs_Arab | | 70 vs 65 | 63 vs 56 | 47 vs 44 | 47 vs 47 | 43 vs 39 | 52 vs 49 | | 72 vs 67 | 66 vs 60 | 47 vs 45 | 45 vs 45 | 44 vs 40 | 53 vs 50 |
| sat_Beng | | 13 vs 5 | 10 vs 2 | 6 vs 2 | 5 vs 3 | 5 vs 1 | 7 vs 2 | | 15 vs 6 | 12 vs 4 | 6 vs 2 | 5 vs 4 | 5 vs 2 | 8 vs 3 |
| sin_Sinh | | 24 vs 22 | 18 vs 13 | 11 vs 8 | 14 vs 14 | 9 vs 5 | 14 vs 12 | | 27 vs 23 | 22 vs 15 | 12 vs 8 | 14 vs 14 | 10 vs 6 | 16 vs 12 |
| snd_Arab | | 33 vs 31 | 26 vs 22 | 16 vs 14 | 19 vs 20 | 13 vs 10 | 21 vs 18 | | 34 vs 32 | 28 vs 24 | 16 vs 14 | 20 vs 20 | 13 vs 10 | 21 vs 19 |
| tam_Taml | | 71 vs 71 | 63 vs 63 | 48 vs 51 | 48 vs 53 | 44 vs 46 | 53 vs 55 | | 73 vs 73 | 66 vs 66 | 49 vs 51 | 47 vs 51 | 45 vs 47 | 54 vs 56 |
| tel_Telu | | 72 vs 72 | 64 vs 64 | 49 vs 52 | 48 vs 52 | 45 vs 47 | 54 vs 56 | | 74 vs 74 | 68 vs 68 | 51 vs 54 | 47 vs 52 | 47 vs 50 | 55 vs 58 |
| tgk_Cyrl | | 62 vs 17 | 52 vs 11 | 37 vs 8 | 38 vs 14 | 33 vs 7 | 42 vs 11 | | 64 vs 16 | 55 vs 11 | 38 vs 7 | 37 vs 13 | 34 vs 6 | 43 vs 10 |
| uig_Arab | | 23 vs 25 | 15 vs 14 | 9 vs 10 | 13 vs 16 | 7 vs 6 | 12 vs 13 | | 23 vs 26 | 17 vs 16 | 9 vs 10 | 12 vs 16 | 7 vs 6 | 13 vs 14 |
| urd_Arab | | 71 vs 71 | 63 vs 64 | 48 vs 51 | 48 vs 52 | 44 vs 47 | 53 vs 56 | | 73 vs 73 | 67 vs 67 | 48 vs 51 | 47 vs 51 | 45 vs 48 | 54 vs 56 |
| uzn_Latn | | 64 vs 60 | 54 vs 50 | 40 vs 39 | 44 vs 45 | 36 vs 35 | 46 vs 45 | | 65 vs 62 | 57 vs 53 | 40 vs 39 | 42 vs 43 | 36 vs 35 | 46 vs 45 |
| Mujadia et al. (2016) | | 51 vs 79 | 45 vs 74 | 36 vs 67 | 36 vs 66 | 32 vs 64 | 39 vs 69 | | 56 vs 80 | 51 vs 76 | 40 vs 69 | 35 vs 66 | 35 vs 66 | 42 vs 70 |
| Overall | | 60 vs 53 | 52 vs 45 | 39 vs 35 | 38 vs 37 | 34 vs 31 | 43 vs 39 | | 63 vs 55 | 56 vs 47 | 40 vs 35 | 38 vs 37 | 36 vs 32 | 44 vs 40 |

Table 11: Language wise performance of wl-coref (Dobrovolskii, 2021) and fast-coref (Toshniwal et al., 2021) in absence of singletons. Notice that performance improves across languages. Indicating that both the models struggles to capture singletons across all languages. Hence advocating the need for coreference resolution models with higher recall in mention detection phase.